\documentclass[10pt,twocolumn,letterpaper]{article}

\usepackage{cvpr}
\usepackage{times}
\usepackage{epsfig}
\usepackage{graphicx}
\usepackage{amsmath}
\usepackage{amssymb}
\usepackage{pifont}
\newcommand{\cmark}{\ding{51}}%
\newcommand{\xmark}{\ding{55}}%
\usepackage{booktabs} 
\usepackage{color}

\usepackage[T1]{fontenc}
\usepackage[utf8]{inputenc}
\usepackage{authblk}
\usepackage[misc]{ifsym}

\usepackage[breaklinks=true,bookmarks=false]{hyperref}

\cvprfinalcopy 

\def\cvprPaperID{****} 

\ifcvprfinal\fi
\begin{document}

\title{DTG-Net: Differentiated Teachers Guided Self-Supervised Video Action Recognition}

\author[1]{Ziming Liu}
\author[1]{Guangyu Gao\thanks{corresponding author}}
\author[2]{A. K. Qin}
\author[1]{Jinyang Li}

\affil[1]{Beijing Institute of Technology, Beijing, China }
\affil[2]{Swinburne University of Technology, Australia}

\maketitle
\thispagestyle{empty}

\begin{abstract}
State-of-the-art video action recognition models with complex network architecture have archived significant improvements, but these models heavily depend on large-scale well-labeled datasets. To reduce such dependency, we propose a self-supervised teacher-student architecture, i.e., the Differentiated Teachers Guided self-supervised Network (DTG-Net). In DTG-Net, except for reducing labeled data dependency by self-supervised learning (SSL), pre-trained action related models are used as teacher guidance providing prior knowledge to alleviate the demand for a large number of unlabeled videos in SSL. Specifically, leveraging the years of effort in action-related tasks, e.g., image classification, image-based action recognition, the DTG-Net learns the self-supervised video representation under various teacher guidance, i.e., those well-trained models of action-related tasks. Meanwhile, the DTG-Net is optimized in the way of contrastive self-supervised learning. When two image sequences are randomly sampled from the same video or different videos as the positive or negative pairs, respectively, they are then sent to the teacher and student networks for feature embedding. After that, the contrastive feature consistency is defined between features embedding of each pair, i.e., consistent for positive pair and inconsistent for negative pairs. Meanwhile, to reflect various teacher tasks' different guidance, we also explore different weighted guidance on teacher tasks. Finally, the DTG-Net is evaluated in two ways: (i) the self-supervised DTG-Net to pre-train the supervised action recognition models with only unlabeled videos; (ii) the supervised DTG-Net to be jointly trained with the supervised action networks in an end-to-end way. The DTG-Net is characterized by fewer training videos, fewer parameters, and lower training costs. Its performance is better than most pre-training methods but also has excellent competitiveness compared to supervised action recognition methods.

\end{abstract}

\section{Introduction}
\label{introduction}
Recently, Video Action Recognition (VidAR) has attracted extensive attention from academia to industry and achieved significant improvement with the robust deep Convolutional Neural Networks (CNNs) architectures [4,29,32]. However, the well-labeled large-scale video dataset is necessary for these State-Of-The-Arts (SOTA) models. Meanwhile, most recent efforts focus on more complex and powerful deep networks with the larger labeled data. In other words, the recognition performance of VidAR methods heavily depends on the large-scale well-labeled video dataset. In practice, it is incredibly time-consuming to annotate these large-scale videos with a temporal dimension. Therefore, there is always a lack of well-labeled video data to train more powerful models in video-based vision tasks, such as VidAR. Consequently, the limitation of a well-labeled dataset further restricts the development and application of powerful action recognition models.

For the challenges of large-scale well-labeled data, Self-Supervised Learning (SSL) provides a promising alternative, where the data itself provides the supervision for learning. SSL methods can be roughly classified into two classes: the generative methods and contrastive methods. For generative methods, it uses the rule-based method to generate labels by raw data itself, e.g., Rotation~\cite{rotation}, Jigsaw Puzzle~\cite{CFNnoroozi2016unsupervised} and Colorization~\cite{vondrick2018tracking}. Meanwhile, the contrastive methods~\cite{CPCoord2018representation, mocohe2019momentum}, also known as Contrastive Self-Supervised Learning, learn representations by contrasting positive and negative examples. So far, there are several studies on SSL based video representation learning, to alleviate the heavy dependency on a large-scale labeled video dataset~\cite{ShuffleandLearn, DPCHan19dpc, 3D-ST-Puzzlekim2019self, 3D-RotNetjing2018selfsupervised, OPNlee2017unsupervised}. Some of these studies belong to generative SSL with different rule-based methods, such as predicting the optical flow~\cite{ActionFlowNet}, predicting the sorting of the video frames~\cite{ShuffleandLearn}. Some others are related to contrastive SSL~\cite{contrastiveVideoAudiokorbar2018cooperative,wang2018contrastiveVideoRepresentation,Temporal_Contrastive_Pretraining}. 

Besides, the SSL is also used for video representation learning and always evaluated on the action recognition task. For example, Dense Predictive Coding (DPC) learned spatio-temporal embedding from video with a curriculum training scheme~\cite{DPCHan19dpc}. Shuffle-and-Learn learned the video representation by predicting the temporal order of the video sequence.~\cite{ShuffleandLearn}. However, although these methods can reduce the dependence on large-scale labeled videos, they fall into the dependence on large-scale unlabeled samples. Namely, robust video representation requires more unlabeled videos for SSL training.

The DistInit~\cite{distinitGirdhar_2019} leveraged easily-obtained image-based vision tasks as teachers to help the action recognition models to obtain a proper parameter initialization. Motivated by this, we realized the contrastive SSL with ideas of knowledge transfer to reduce the dependency on large-scale unlabeled data in SSL. Specifically, leveraging the years of effort in action-related vision tasks, such as image classification and still-image-based action recognition, we propose a self-supervised teacher-student architecture, i.e., the Differentiated Teachers Guided self-supervised Network (DTG-Net) for VidAR.

The DTG-Net utilizes the contrastive self-supervised learning to reduce the dependency on large-scale well-labeled data, and alleviate the dependency on large-scale unlabeled data with guidance of the prior knowledge from action-related vision tasks. 

The DTG-Net is optimized by the \textit{contrastive feature consistency}. The contrastive feature consistency is based on \textit{contrastive pair}, which is the input of DTG-Net. These contrastive pairs conclude the positive pairs and the negative pairs. The positive pair is constructed with two sparse image sequences randomly sampled from the same video. In contrast, those pairs constructed with different videos are defined as negative pairs. After that, the contrastive feature consistency is defined as the feature consistency for positive pair and inconsistency for negative pairs. Meanwhile, the student network's output feature is called \textit{anchor feature}, while that of the teacher network is called \textit{guidance feature}. The goal of the DTG-Net is to make each sample's feature away from others in representation space. Namely, the anchor has pushed closer to the guidance feature of positive pairs but away from the guidance feature of negative pairs.

As a self-supervised teacher-student architecture, it is somehow subjectively biased to decide which task is involved in as the teacher task. To reduce this bias, differentiate teachers guidance is adopted in DTG-Net. Various teacher tasks have different relations to the action recognition task (student). Thus, we explore three different methods to re-weight these multiple teachers' guidance, including one offline method and two online methods. The offline one is to obtain the weights of different teachers based on the accuracy results of DTG-Net with every single teacher, which is obtained in advance. The last two methods compute different weights for different training samples. The second method computes the weights with the similarity of the anchor feature and the guidance feature in the positive pair. The third method obtains the weights based on the ranking of the positive pair's similarity in all of the contrastive pairs.


The SSL based methods are always used for unsupervised pre-training and should be verified in downstream tasks, e.g., VidAR. Therefore, we evaluate the DTG-Net on downsteam task of VidAR in two ways, (i) self-supervised DTG-Net: used to pre-train the networks of downstream tasks with only unlabeled videos; (ii) supervised DTG-Net: jointly learning with network of downstream task end to end. The evaluation of self-supervised DTG-Net is just following the traditional evaluation manner for SSL. The supervised DTG-Net, which combine the DTG-Net with a network of the downstream task, improve the generalization ability of the network in the downstream task, and further reduce the large-scale data dependency. In detail, in supervised DTG-Net, the contrastive SSL can separate different videos in the feature space, especially reducing category overlapping of different actions in the feature space. Namely, supervised DTG-Net can provide better inter-class variance between different action categories with existing networks.

Finally, the\textbf{ main contributions} of this work can be summarized as follows:
\begin{enumerate}
\item We propose a self-supervised teacher-student architecture, the DTG-Net, to learn video representation with easily-obtained unlabeled data and related pre-trained models. The teacher tasks in DTG-Net provide prior knowledge to accelerate the self-supervised training further and reduce dependency on unlabeled data. 
\item We firstly realize the video related contrastive self-supervised learning by contrastive feature consistency, which is defined on randomly sampling two different image sequences from the same video as positive pairs, and that from two different videos as negative pairs.
\item In evaluation, we propose the supervised DTG-Net by combining the DTG-Net with a supervised action network, which increases the generalization ability of the action recognition network. 
\item With above advantages, DTG-Net achieves state of the art performance with less video data and less training cost.
\end{enumerate}

\section{Related works}
\label{sec:related_works}

DTG-Net is self-supervised teacher-student architecture. Also, we design the supervised DTG-Net for modifying the supervised action recognition models. Therefore, we review the recent related works about self-supervised learning and video action recognition. 

\subsection{Self-supervised Learning}

\subsubsection{\textbf{Context based self-supervised learning}}
There are many self-supervised tasks constructed with the context of the image data. Carl et al.~\cite{Jigsaw1doersch2015unsupervised} proposed Jigsaw to construct image spatial context as the supervisory signal for self-supervised learning. Noroozi et al.~\cite{CFNnoroozi2016unsupervised} extended the first work, introducing the context-free network (CFN) to solve the Jigsaw puzzles self-supervised task. Deepak et al.\cite{contextinpaitingpathak2016context} performed self-supervised learning by context-based pixel prediction. 
Zhang et al.~\cite{zhang2016colorful} solved the task of hallucinating a plausible color version of the photograph with grayscale photography as input. 
Besides directly defining a self-supervised learning task, there are also more and more works exploring downstream task-related self-supervised learning. Lee et al.~\cite{lee2019rethinkingdataaugmentation} proposed to learn the joint distribution of the original labels and self-supervised labels of augmented samples. A similar idea is also applied to a few shot learning~\cite{fewshotselfsupervisedGidaris_2019_ICCV}. For DTG-Net, we also explore action recognition task-related self-supervised learning, i.e., supervised DTG-Net.

\subsubsection{\textbf{Temporal base Self-supervised Learning}}
For the video data, most self-supervised learning tasks are related to temporal constrains. 
Time-Contrastive Network (TCN)~\cite{timecontrastivenetsermanet2017timecontrastive} achieved self-supervised learning from unlabeled videos recorded from multiple viewpoints. 
Wang et al.~\cite{wangxiaolong2015videounsupervised} proposed to use visual tracking to provide supervision on unlabeled video data. Shuffle and Learn~\cite{ShuffleandLearn} defined a self-supervised task that learns whether the image sequences of the same video are in the correct temporal order. OPN~\cite{OPNlee2017unsupervised}, 3D-RotNet~\cite{3D-RotNetjing2018selfsupervised}, 3D-ST-Puzzle~\cite{3D-ST-Puzzlekim2019self} also define self-supervised learning tasks based on different video properties. Different from these works, DTG-Net adopts a more simple way to realize the video-based self-supervised learning, with the instance discrimination contrastive learning~\cite{mocohe2019momentum, instance_discriminationwu2018unsupervised} and under the guidance of action-related teacher tasks. 

\subsubsection{\textbf{Contrastive based Self-supervised Learning}}
Contrastive learning has become one of the most common and basic methods for self-supervised learning.
DeepInfoMax~\cite{deepinfomaxhjelm2018learning} defines a contrastive task by recognizing if a pair of global and local features are from the same image. CPC~\cite{CPCoord2018representation} is designed for sequence data,  which learns representations by predicting the future in latent space using auto regressive models.  AMDIM~\cite{AMDIMbachman2019amdim} and CMC~\cite{CMCtian2019contrastive} learn invariance of data augments and different views to achieve self-supervised learning.  Contrastive Learning has been used to solve different tasks. DPC~\cite{DPCHan19dpc} learns a dense encoding of spatio-temporal blocks by recurrently predicting future representations to learn video representation. CRD~\cite{CRDtian2019contrastive} applies contrastive learning to help knowledge transfer and knowledge distillation. 
The recent contrastive learning method, MoCo~\cite{mocohe2019momentum}, builded a dynamic dictionary with a queue and a moving-averaged encoder to solve the training problem of previous Instance discrimination contrastive learning~\cite{instance_discriminationwu2018unsupervised}. The simCLR is another simple contrastive learning method~\cite{simCLRchen2020simple}, but it depends on expensive training costs.   DTG-Net also adopted the contrastive learning design of the MoCo~\cite{mocohe2019momentum}, but we introduce the pre-trained teacher guidance to accelerate the self-supervised training process. 



\subsection{Supervised Action Recognition}

The deep learning based action recognition models benefit from the deep CNNs, the large-scale well-labeled data, as well as the good parameter initialization. To better understand the student network of DTG-Net, we review the supervised action recognition methods in this subsection. 

Most supervised action recognition methods could be cluster into three types, i.e., CNN-based model, multi-modality model, RNN-based model. These three methods could be used together, but the CNN-based model is the most commonly used architecture~\cite{I3D, C3D}. Therefore, we use the CNN-based model (TSN and 3D-ResNet) as the student network of DTG-Net.   
TSN~\cite{TSN}, TRN~\cite{TRN} are 2D CNN-based methods, which learn the video representation considering the structure information of videos. I3D, C3D, SlowFast Net are the most commonly used 3D CNN-based action recognition models, which achieves the state of the art results in most public data sets~\cite{I3D, C3D,feichtenhofer2019slowfast}. Tiny video network~\cite{tinyvideonet} is a light-weight CNN model, designed with neural architecture search (NAS). Dist Init~\cite{distinitGirdhar_2019} is also a teacher-student action network, designed to initialize the parameters of the various architecture with image task's pre-trained models. DTG-Net is motivated by the Dist Init, utilizing easily-obtained image task's pre-trained networks to accelerate the contrastive self-supervised learning. But the optimization goal of DTG-Net is different from Dist Init, as explained in Section \ref{sec:teacher_guidance}. 

\begin{figure*}[t]
    \centering
    \includegraphics[width= \textwidth]{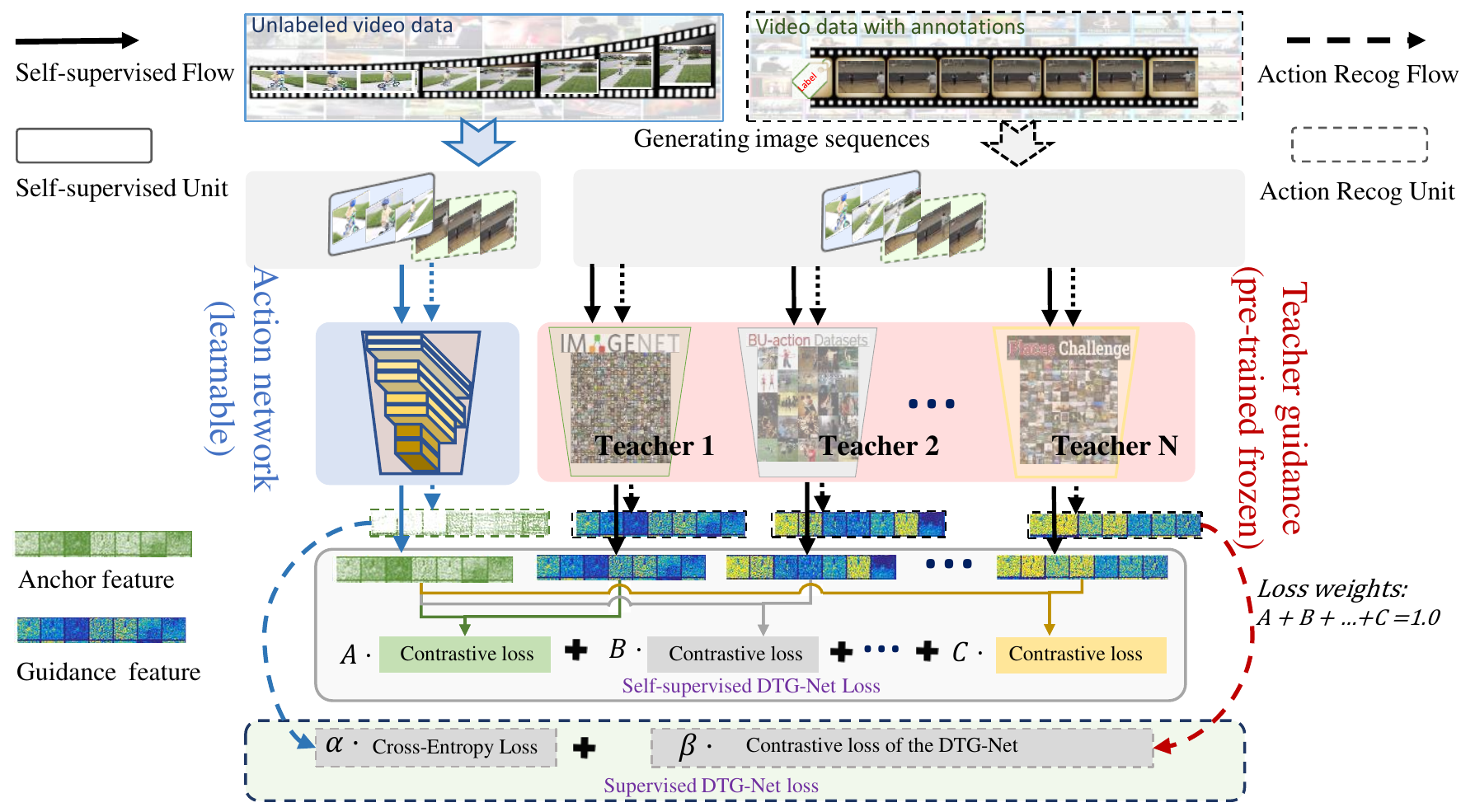}
    \caption{The process of the whole DTG-Net, with self-supervised task and action recognition (downstream) task.}
    \label{fig:DTG-Net}
\end{figure*}

\section{Method}
\label{methods}




\subsection{Overall about DTG-Net}

DTG-Net is a self-supervised teacher-student architecture that aims to solve the data dependency problem and reduce the training cost. The DTG-Net is trained in a self-supervised way, i.e., contrastive self-supervised learning. The optimization goal of this student network is to make the anchor feature (action recognition feature) closer to guidance features from the same sample, but away from those guidance features from different samples. The anchor feature is the output of the student network. All guidance features are obtained by multiple differentiate pre-trained teacher networks, each of which corresponds to an easily-obtained action-related task. 



\subsection{Contrastive Pairs}
There is a video $v = \mathbb{R}^{T\times H \times W}$, where $T, H, W$ denote temporal range, frame height, frame width separately. Two different image sequences $\{v_{anchor}, v_{guidance}\}$ are obtained by random frame sampling from the same original video $v$. The two inputs correspond to two different types of networks in Figure \ref{fig:DTG-Net}, i.e. frozen pre-trained public teachers $f_{guidance}(\cdot)$ and student network $f_{anchor}(\cdot)$~\cite{mocohe2019momentum,instance_discriminationwu2018unsupervised}.
Following common practice, the input of any network is a sparse image sequence generated from the original video by data augment operations. The data augment operations include dividing a video into $T$ segments, random sampling one frame from each segment, random cropping the $T$ images of a sparse image sequence. The two inputs of DTG-Net ($\{v_{anchor}, v_{guidance}\}$) are obtained with the above data augment operations. Each of them contains parts of information of the same video. 

To optimize the student network with contrastive self-supervised learning, contrastive pairs are built with $\{v_{anchor},  v_{guidance}\}$. The $v_{anchor, i}$ and $v_{guidance, i}$ from the same video form a positive pair, while the $v_{anchor, i}$ and $v_{guidance, j}, j \in Q$ from different videos form an negative pair. $Q$ is a feature set that is a queue and updated with the training iterators~\cite{mocohe2019momentum}. As the optimization goal of the DTG-Net, the positive pair should be more similar, while the negative pairs should be different.

\subsection{Teacher Guidance}
\label{sec:teacher_guidance}
As mentioned above, the DTG-Net is a teacher-student structure. The student network is the target network to be optimized. The teacher network performs as the guidance for optimizing the target network. 
The design of the teacher guidance is the most crucial part of the DTG-Net, which is motivated by the Dist Init method ~\cite{distinitGirdhar_2019}. 
The teacher network encoders the input image sequence into a feature representation space $R_{guidance}$ of prior knowledge. Those anchor features of student network form another feature representation $R_{anchor}$. Those features from $R_{guidance}$  help to refine those anchor features of $R_{anchor}$. 

The teacher network introduces prior knowledge from action-related tasks. For example, image-based action recognition is easier, and whose pre-trained model can be obtained at a low cost. This prior knowledge performs as teacher guidance to build a feature representation space $R_{guidance}$.

Because of the prior knowledge, the training of DTG is more stable, and DTG accelerates self-supervised learning. 
The teacher network is pre-trained on other tasks, and its parameters are frozen in DTG. The fixed parameters promise the training of DTG is stable because only one student network is optimized. This solves the training difficulty of previous simple contrastive learning~\cite{mocohe2019momentum}. Also, the introduction of prior knowledge makes the DTG-Net require fewer data and be optimized faster.

\begin{figure}
    \centering
    \includegraphics[width=0.45\textwidth]{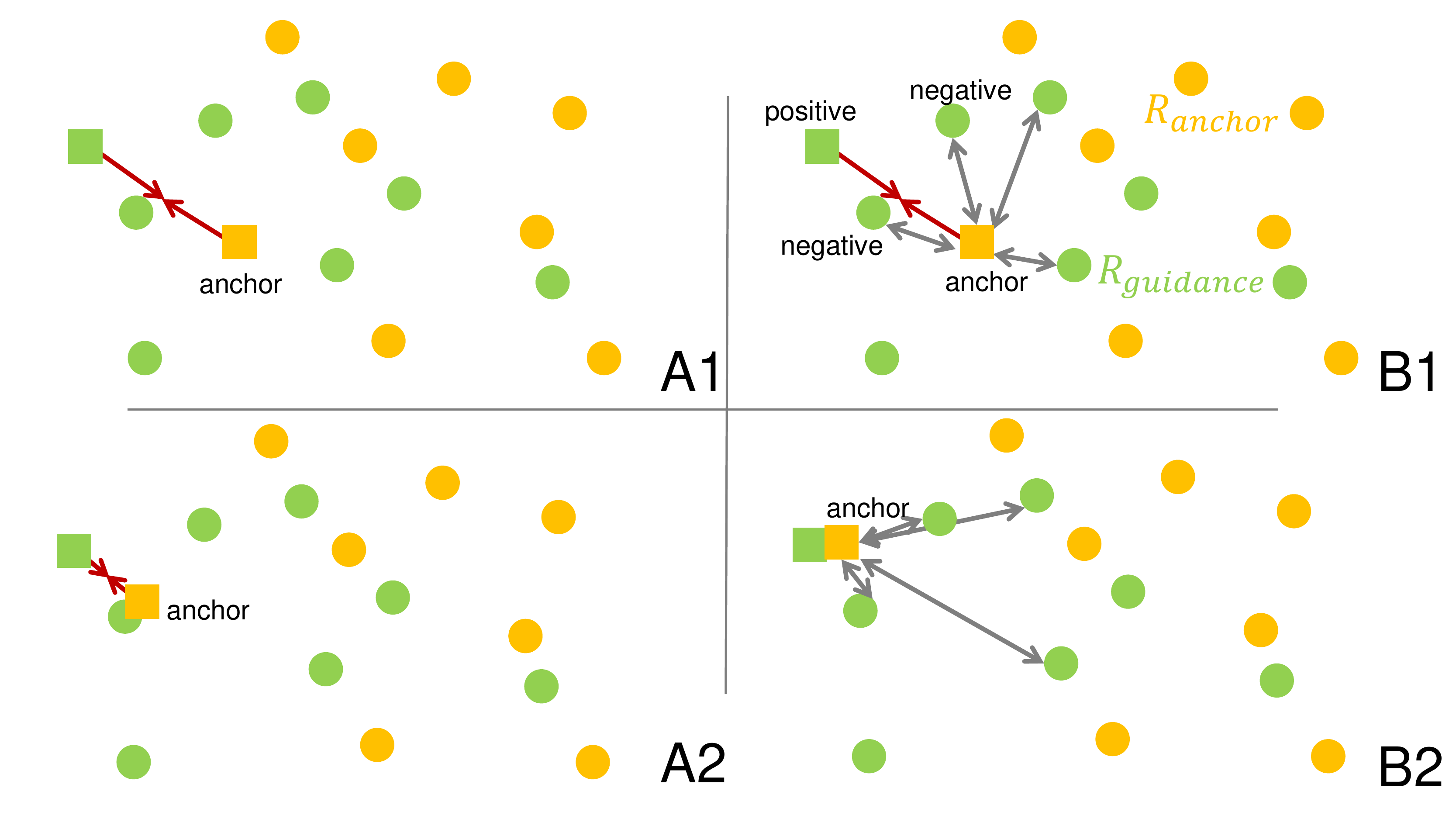}
    \caption{The comparison of DTG and Dist Init. $A1$ to $A2$ shows the distribution of Dist Init, $B1$ to $B2$ shows the distribution of DTG. The rectangle features are from the same sample, and those circle features are from other samples. Those orange features represent $R_{anchor}$, those green features represent $R_{guidance}$.}
    \label{fig:difference_distinit}
\end{figure}

Different from the Dist Init\cite{distinitGirdhar_2019}, which also introduces the prior knowledge of image-based tasks, the Dist Init fits the anchor feature representation $R_{anchor}$ into the teacher feature representation $R_{guidance}$ with L1 loss or cross-entropy loss. In DTG-Net, the teacher feature representation $R_{guidance}$ has two forces to build the $R_{anchor}$, making the anchor closer to that guidance of the same video and away from that guidance of different videos. 
As Figure \ref{fig:difference_distinit} shown, the Dist Init only fits each sample to the corresponding teacher feature, the anchor feature may be closer to other samples. But the DTG considers multiple videos, and there are two forces to push an anchor feature in the right place in $R_{anchor}$.

\subsubsection{\textbf{Differentiated Teacher Guidance}}
\label{the_weighted_guidance}
There could be multiple teacher networks for the DTG task. Each teacher network corresponds to a specific action-related vision task. In this subsection, we explore to use $N$ teachers from $N$ tasks to guide the building of representation space $R_{anchor}$. Each teacher network builds a prior knowledge representation space $R_{guidance, k}, k \in [0,N)$.

\begin{table*}[t]
  \centering
  \setlength{\tabcolsep}{4.7mm}
  \begin{tabular}[]{c|c|c|c|c}
       \toprule
       Model&Backbone(\#param)&SSLDataset&UCF101&HMDB51\\
       \midrule
       Random init~\cite{DPCHan19dpc}&3D-ResNet18(\#33.4M)&-&46.5&17.1\\
       ImageNet init~\cite{twostream}&VGG-M-2048(\#25.4M)&-&73.0&40.15\\
       \hline
       Random init&3D-ResNet18(\#33.4M)&-&50.79&-\\
       \dag ImageNet init&TSN-ResNet18(\#33.4M)&-&73.4&-\\
       \dag ImageNet init&3D-ResNet18(\#31.8M)&&79.3&43.6\\
       BU init&3D-ResNet18(\#31.8M)&&85&46.4\\
       \midrule
       \raggedright{(2016)Shuffle\&Learn}&CaffeNet(\#58.3M)&UCF/HMDB&50.2&18.1\\
       (2017)OPN&VGG-M-2048(\#25.4M)&UCG/HMDB&59.8&23.8\\
       (2019)DPC&3D-ResNet18(\#14.2M)&UCF&60.6&-\\
       Ours(ImageNet-guidance)&TSN-ResNet18(\#11.2M)&UCF&70.8&32.6\\
       Ours(Kinetics-guidance)&3D-ResNet18(\#31.8M)&UCF/HMDB&71.3&41.1\\
       \dag Ours(bu+stanford-guidance)&3D-ResNet18(\#31.8M)&UCF/HMDB&\textbf{85.6}&\textbf{49.9}\\
       \midrule
       (2018)3D-RotNet&3D-ResNet18(\#33.6M)&Kinetics400&62.9&33.7\\
       (2019)3D-ST-Puzzle&3D-ResNet18(\#33.6M)&Kinetics400&63.9&33.7\\
       (2019)DPC&3D-ResNet18(\#14.2M)&Kinetics400&68.2&34.5\\
       (2019)Dist init&R(2+1)D18(-)&Kinetics400&-&40.3 \\
       Ours(ImageNet-guidance)&TSN-ResNet18(\#11.2M)&Mini-Kinetics200&67.0&-\\
       Ours(ImageNet-guidance)&TSN-ResNet18(\#11.2M)&Kinetics400&69.1&-\\
       \bottomrule
  \end{tabular}
  \vspace{0.5cm}
  \caption{The comparison with video-based self-supervised methods. \dag\ means the action recognition network is initialized with ImageNet pre-trained 2D CNN. \textit{SSLDataset} is the data set used for the self-supervised task, and then the evaluation results(top1 accuracy) of these models on UCF101 and HMDB51 are reported.}
  \label{tab:sota}
\end{table*}
Specifically, we use still-image-based action recognition tasks (BU101, Stanford40), object recognition (ImageNet), and scene recognition (Places365) as teacher guidance tasks. Although prior knowledge of these tasks could be used for guiding the target network, different tasks have different relation, as mentioned in Task Similarity Tree~\cite{Taskonomy}. The relation difference can be shown as different loss weights of different teachers. 

To balance the multiple guidance tasks to achieve a better guidance effect, we assign different weights to different guidance tasks. Considering $N$ teacher guidance tasks, the fused teacher guidance $R_{guidance}$ is the weighted average of them, as shown in Eq. \ref{w_setting}. We try three different weighting methods in this paper, as described in Section \ref{sec:ablation_study_differentiatedTeacherGuidance}.

\begin{equation}
    \label{w_setting}
     R_{guidance}=\sum^{N-1}_{k=0}w_{k} \cdot R_{guidance, k}\\, where
     \sum^{N-1}_{k=0}w_{K}=1
\end{equation}

\subsubsection{\textbf{Advantages of DTG}}
The DTG-Net has two low-cost characteristics, i.e., less training data and easily-obtained pre-trained teachers. And it overcomes the training difficulties of the simple contrastive self-supervised learning.

\subsection{Contrastive Loss in DTG-Net}
\label{sec:contrastive_loss}
As the last section mentioned, there are two forces to build the $R_{anchor}$, the two forces are performed with the contrastive loss, coming from positive pairs and negative pairs. To construct the negative pairs, there is a queue $Q$ of size $K$, which is continually updated.  This maintained feature set is parts of the $R_{guidance}$. The size of $K$ is smaller than the total training samples of $N$. Because the inputs of the DTG-Net are sparse image sequences generated with random data augment operations, the feature set $Q$ is not constant, and the features in $Q$ are updating with an iterator.For each new video $v_{\alpha}$, positive pair is constructed with the guidance feature and the anchor feature of the same video. The negative pairs are constructed with the feature set $Q$ and the anchor feature of this video.

The contrastive loss is computed following the common practice of Instance Discrimination Contrastive Learning~\cite{instance_discriminationwu2018unsupervised,mocohe2019momentum}. A common contrastive loss function, i.e., InfoNCE, is used here. The similarity of any two features (no matter positive pair or negative pair) is defined by dot product as in~\cite{mocohe2019momentum}, $Similarity = a_{\alpha}\cdot g_{i}$. The goal of the DTG is to minimize the Eq. \ref{inforNCE}, i.e. maximize the similarity of $a_{\alpha}$ and $g_{\alpha}$ of a same video. The $a$ and $g$ represent the anchor feature and the guidance feature. 
\begin{equation}
    \label{inforNCE}
    L_{contrast}=-log \frac{exp(a_{\alpha}\cdot g_{\alpha}/\tau)}{\sum_{i=0}^{K} exp(a_{ \alpha}\cdot g_{i}/\tau)}
\end{equation}

The $\tau$ is a temperature hyper-parameter to scale the value of the vector similarity ~\cite{instance_discriminationwu2018unsupervised}. There are total $(K+1)$ similarity values, including $1$ positive similarity and $K$ negative similarities. This form of InfoNCE comes from~\cite{mocohe2019momentum}, which is efficient and straightforward. There are also other forms of the InfoNCE loss function ~\cite{CPCoord2018representation}, but the forms of the loss function are not the focus of this paper.

\section{Experimental Results}
In this section, we evaluate the DTG-Net on VidAR in two ways, (i)self-supervised DTG-Net: used to pre-train the networks of down-stream tasks with only unlabeled videos; (ii) supervised DTG-Net: jointly learning with a network of downstream task end to end.

\subsection{Datasets}
There are three data sets used for these experiments, including Kinetics400, UCF101 and HMDB51. The Mini Kinetics200 generated from Kinetics400 are also used for the self-supervised task of the DTG-Net. 
\begin{itemize}
  \item \textbf{Kinetics400}, Kinetics400 is a large-scale action recognition benchmark data set with more than 300k videos with 400 action categories. These videos mainly come from the YouTube video website.~\cite{S3Dxie2018rethinking}.
  \item  \textbf{HMDB51}, HMDB51 is the first widely used video action recognition data set with 51 action labels. It has been an essential benchmark for most action recognition models. But the size of this data set is much smaller than the recently released video data sets. 
  \item  \textbf{UCF101}, UCF101 is another popular and important action recognition benchmark data set. There are 101 action categories and about 10k videos in this data set. This data set is larger than the HMDB51, but easier to be recognized compared with the HMDB51.
\end{itemize}

\subsection{The Training Detail}
\label{trainingdetail}
We perform all experiments on Nvidia 2080Ti and 1080Ti GPUs. For the self-supervised learning with the DTG-Net in UCF or HMDB data set, experiments are conducted with an initial learning rate of 0.1. SGD optimizer is used with a momentum of 0.9 and a weight decay of 0.0005. We train these models for 600 epochs, with the learning rate multiplied by 0.1 at the 300 and 500 epochs. For the self-supervised task in Kinetics400 or Mini-Kinetics200, the initial learning rate is also 0.1, and it is optimized for 200 epochs with the learning rate multiplied by 0.1 at 100 and 150 epochs on 8 GPUs.

For the supervised DTG-Net, the ablation experiments are conducted with 0.1 initial learning rate, using the SGD optimizer. The learning rate multiplied by 0.1 at 30 and 60 epochs.  All of the models are optimized for 80 epochs in UCF101 and HMDB51.

\subsection{Self-supervised DTG-Net}
\label{sec:self-supervised_DTG-NET}
In the self-supervised DTG-Net, the contrastive feature consistency is defined on the construction of contrastive pairs by different frame sampling strategies. In self-supervised DTG-Net, we froze the teacher models to train the student model (a VidAR network). This VidAR network is used as a pre-trained network for the downstream task of supervised VidAR on the popular dataset, e.g., UCF and HMDB. Generally, the VidAR networks should be the same in self-supervised learning and downstream task but could be any supervised action recognition networks. 

\subsubsection{\textbf{Comparison with SOTA}}

We compare the self-supervised DTG-Net and other State-Of-The-Art video-based self-supervised learning methods, by performance in the downstream VidAR on UCF101 and HMDB51. Specifically, we make the comparison by conducting different Self-Supervised Learning (SSL) on both small-scale datasets (UCF101, HMDB51) and large-scale datasets (Kinetics400, Mini-Kinetics 200). It is worthy to note that the self-supervised task does not use any labels. Finally, for evaluation in the downstream VidAR task, the DTG-Net achieves the best performance when trained on the UCF101 and HMDB51, and also surpass some other methods when trained on Kinetics400, shown in Tab~\ref{tab:sota}. 

Furthermore, to make a more fair comparison with the traditional ImageNet based pre-trained models, we compared the VidAR performance of ImageNet pre-trained 3D-ResNet18 network with or without the DTG-Net self-supervised task.
The DTG-Net here uses image action recognition on BU101 and Stanford40 data sets (two small-scale still-image-based action recognition data sets) as the teacher guidance. The results show that the DTG-Net achieves more than $6.3\%$ top1 accuracy improvement compared with the baseline and suppresses the network, which is initialized with the BU101 pre-trained model. Besides, we also train the DTG-Net on Kinetics400, and Table \ref{tab:sota} shows that DTG-Net suppresses the state-of-the-art methods with fewer parameters.

\subsubsection{\textbf{Ablation Study: the inputs of DTG-Net}}

\begin{table}[h!]
  \centering
  \setlength{\tabcolsep}{1.2mm}
  \begin{tabular}[]{c|c|c|c|c}
       \toprule
       Teacher&Student&Overlap&$acc@1$&$acc@5$\\
       \midrule
       img->TSN&img->TSN&\cmark&57.73&80.07 \\
       img->TSN&seq->3D-ResNet&\cmark&58.02&79.62 \\
       seq->TSN&seq->TSN&\cmark&\textbf{58.76}&80.84 \\
       seq->TSN&seq->TSN&\xmark&57.65&79.73 \\
       \bottomrule
  \end{tabular}
  \vspace{0.5cm}
  \caption{The ablation study of the input of the DTG-Net. The self-supervised task and the evaluation are both conducted on UCF 101 split1. The ``Overlap" refers to if there is a temporal overlapping for multiple image sequences from the same video. \textit{img} is any frame in a video. \textit{seq} is the sparse image sequence generated from a video.}
  \label{tab:theinput}
\end{table}

Since the teacher tasks in DTG-Net are mostly the 2D vision tasks, we test different inputs of DTG-Net with video dataset, such as an image sequence or a single image from a video. Since we already know that a longer image sequence is better for supervised action recognition models, the input type is significant for action recognition task. In this experiment, we compare four different types of inputs for DTG-Net, i.e. 1) sampling two different images from the same video, 2) sampling one image and one image sequence from the same video, 3) sampling two different image sequences with temporal overlapping, 3) sampling two different image sequences without temporal overlapping. 

As shown in Table~\ref{tab:theinput}, image sequences are more suitable as inputs for DTG-Net compared with the single frame. This result suggests that the DTG-Net also benefits from a \textit{dense input}, in which more video information is involved. Besides, since the contrastive loss in DTG-Net requires different inputs of DTG-Net, we evaluate DTG-Net by two different image sequences sampling strategies, i.e., sampling the sequences with or without temporal overlapping. The result suggests that sampling image sequences with temporal overlapping achieves satisfactory performance. It could be that two sequences generated in this way have more similar information distribution, which is suitable for DTG.

\begin{table}[h!]
  \centering
  \setlength{\tabcolsep}{1.8mm}
  \begin{tabular}[]{c|c|c|c|c}
       \toprule
        TeacherTask&Weights&DataSize&$acc@1$&$acc@5$\\
       \midrule
       ImageNet~\cite{ImageNet}&&1.28M&59.98&82.47\\
       Places~\cite{zhou2017placesdataset}&&1.8M&49.96&72.54 \\
       BU~\cite{BU101datasetma2017less}&&23.8K&62.01&84.19 \\
       Stanford~\cite{stanford40}&&9.5K&61.83&83.48 \\
       \hline
       ImageNet+Places&\xmark&3.08M&54.77&77.74 \\
       
       BU+Standford&\xmark&33.3K&62.65&83.61 \\
       \hline
       All&\xmark&3.46M&60.45&82.84  \\
       All&offline&3.46M&63.28&84.38  \\
       All&online1&3.46M&61.59&82.47  \\
       All&online2&3.46M&59.48&82.18\\
       \bottomrule
  \end{tabular}
  \vspace{0.5cm}
  \caption{Results of self-supervised DTG-Net with differentiate teacher guidance. The evaluation is conducted on UCF101 split1. The \textit{offline} means the weights of teacher guidance are determined by the accuracy results of DTG-Net with this single teacher. The \textit{online1} means the weights are determined by the dot product similarity of the positive pairs of multiple teacher tasks. The \textit{online2} means the weights are determined by the ranking of the similarity of positive pair in all of the contrastive pairs.}
  \label{tab:multi-data}
  \vspace{-0.5cm}
\end{table}

\subsubsection{\textbf{Ablation Study: Differentiated Teacher Guidance}}
\label{sec:ablation_study_differentiatedTeacherGuidance}
To reflect the various influences of different teachers, we also explore the effect of diverse guidance weights assigned on teacher tasks. We first make a comparison of the results of DTG-Nets with different single teacher task. There are four VidAR related tasks considered here, i.e., the image classifications on ImageNet, scene recognition on Places, image action recognition on BU, and Stanford40. Intuitively, static image-based action recognition is the most related tasks with VidAR. 

Secondly, we investigate three weighting ways for multiple teachers with various guidance, namely, weighing each contrastive loss between a teacher and the student task. These three weighting ways include one offline way and two online ways. (1) The offline way performs DTG-Nets with each teacher task separately to obtain their accuracy results, namely, the softmax output on the top1 accuracy. Then these accuracy values are used as the weights, such as ($0.067, 3.0\times 10^{-6}, 0.51, 0.43$) in this experiment. (2) The first online way outputs different the loss weights for each training sample. As the goal of the DTG-Net is the feature consistency between the anchor feature and guidance feature, we use the dot product similarity of the positive pair as the weights of corresponding contrastive loss. (3) Similar to the online1 method, the online2 method also gives different loss weights for different training samples. The difference is online2 method determines the loss weights by the ranking of the similarity of the positive pair in all of the contrastive pairs.

The results in Table \ref{tab:multi-data} suggest that the balance of multiple teacher guidance is essential, while there is a $2.83\%$ top1 accuracy improvement by re-weight the contrastive losses of DTG-Net. In addition, the comparison of three re-weight methods suggests that the offline method is better than online methods. For online methods, the value of the feature similarity is more suitable than the ranking of feature similarity. Also, we visualize the training process of the online1 method, and the result is similar to the offline method ($BU > Stanford > ImageNet > Places$). Although these online methods are not as good as the offline method, their training cost is lower than the offline method. Therefore, the balance between the accuracy and training cost need to be considered. 
\begin{figure}
    \centering
    \includegraphics[width= 0.5\textwidth]{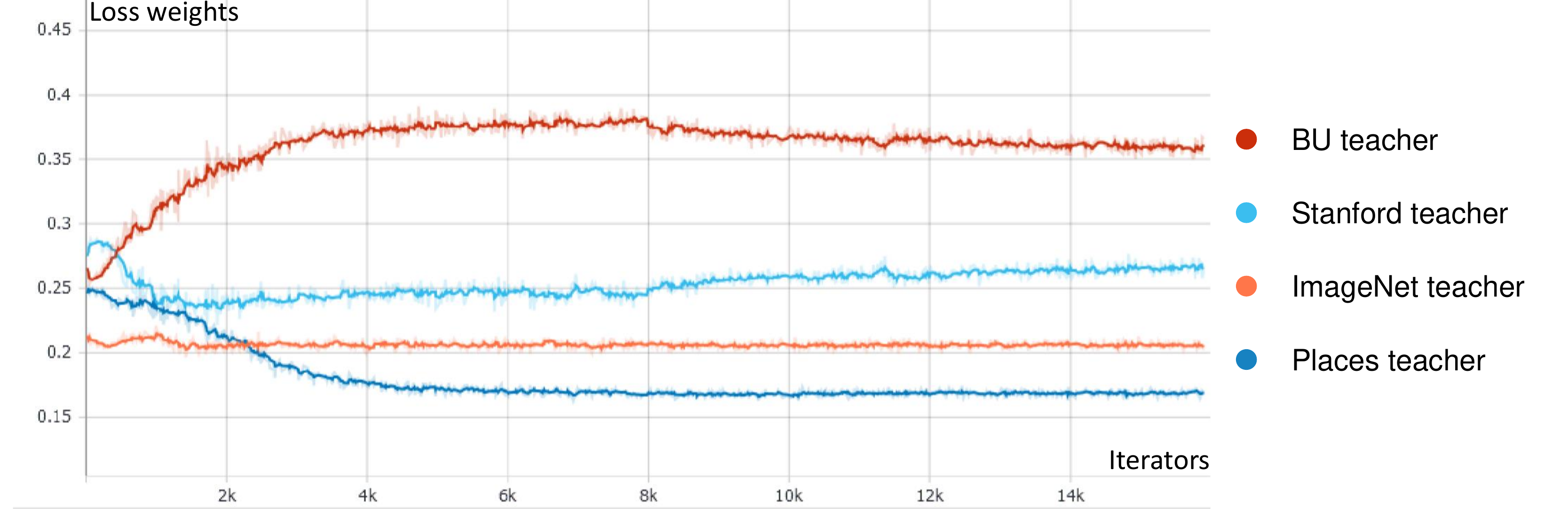}
    \caption{The training process of the online1 method.}
    \label{fig:vis_weight}
\end{figure}

\subsection{Supervised DTG-Net}

\begin{figure}[h!]
    \centering
    \includegraphics[width=0.46\textwidth]{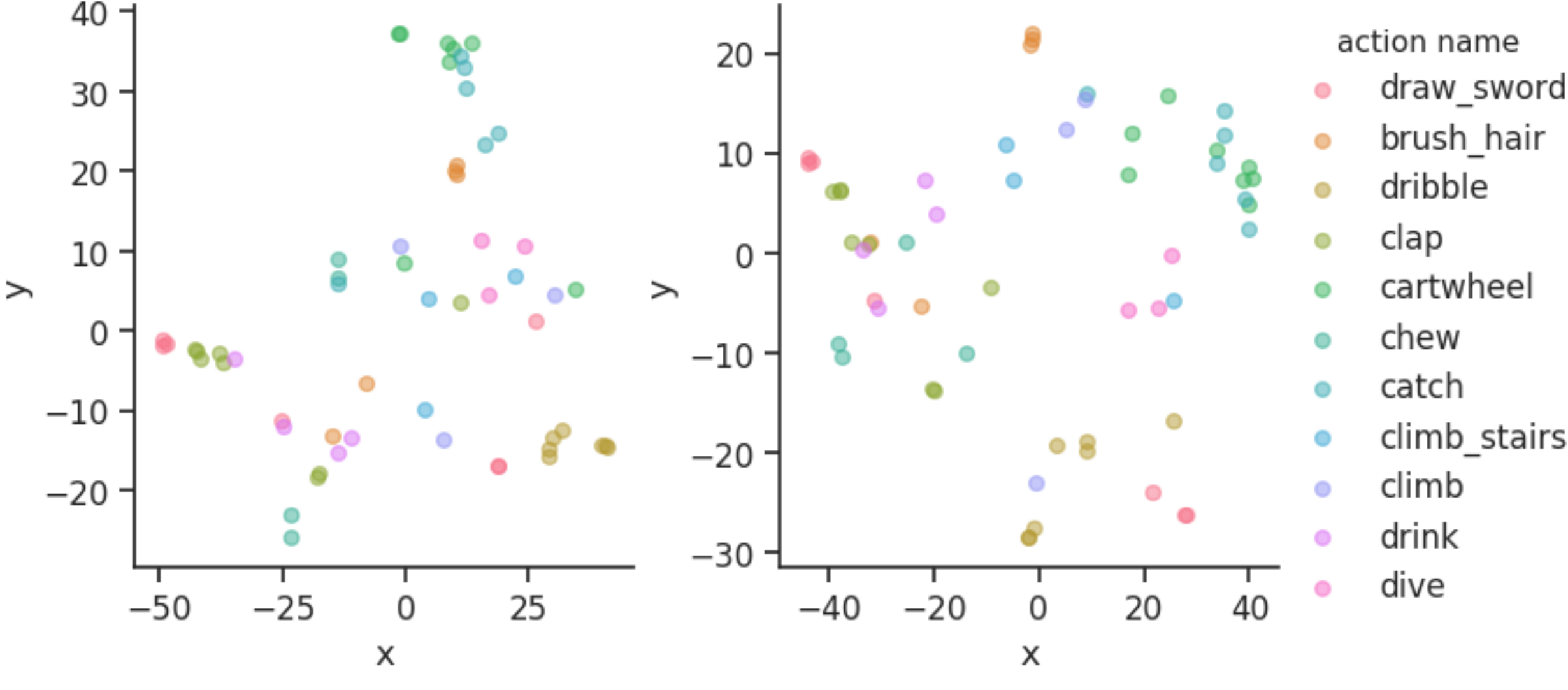}
    \caption{The visualization of the feature representation space $R_{anchor}$ for the last experiment in Table \ref{tab:ablationstudySupervisedDTG}. (1):  \textit{supervised DTG-Net} (2): \textit{supervised action network without DTG-Net}. The features came from the last layer's outputs of the action network and visualized with the t-SNE technique~\cite{tsnemaaten2008visualizing}. To be convincing, we directly visualize those features from the \textit{top10 classes in the HMDB51 official annotation files}.}
    \label{fig:visualization_supervised_dtg-net}
\end{figure}
\label{sec:supervised_DTG-Net}

The \textit{supervised DTG-Net} is just a supervised video action recognition model, which used the labeled videos to train the DTG-Net. Specifically, the raw videos are sent to the DTG-Net to generate the contrastive loss, and videos with labels are also sent to the VidAR network (the student network in DTG-Net) in a supervised way for the cross-entropy loss. After that, these two losses are combined for joint learning, as shown in Figure \ref{fig:DTG-Net}.

\subsubsection{\textbf{The experiment procedure}}
Supervised DTG-Net not only performs the supervised action recognition training but also keep the self-supervised structure of DTG-Net. In a word, the supervised DTG-Net is the combination of supervised action recognition and self-supervised learning technique. Specifically, the student network in supervised DTG-Net could be supervised action recognition models, e.g., TSN, 3D-ResNet, I3D, C3D~\cite{TSNWang_2016, C3D, I3D}. The teacher networks of supervised DTG-Net keep the same with that of the self-supervised learning in Section \ref{sec:self-supervised_DTG-NET}. Therefore, there are two losses in the supervised DTG-Net, i.e., contrastive loss and cross-entropy loss, whose weights are $\alpha=0.1$ and $\beta=1.0$ separately. The goal of supervised DTG-Net is to optimize the joint loss of $L_{joint}$ in Eq. \ref{joint_loss}.
 
\begin{equation}
\begin{aligned}
    L_{joint} &= \alpha \cdot L_{CT} + \beta \cdot L_{CE} 
    \\ &= -\alpha \cdot log \frac{exp(a_{\alpha}\cdot g_{\alpha}/\tau)}{\sum_{i=0}^{K} exp(a_{ \alpha}\cdot g_{i}/\tau)} -\beta \cdot log \frac{exp(l_{+})}{\sum_{j=0}^{N} exp(l_{j})} 
\end{aligned}
\label{joint_loss}
\end{equation}
The $L_{CT}$ and $L_{CE}$ refer to contrastive loss of DTG-Net and cross-entropy loss of supervised action recognition. 

As mentioned before, the $L_{CL}$ forces each video stay away from each other in the feature space. At the same time, $L_{CE}$ put the videos of the same category to be closer and videos of different categories to be away from each other. The $L_{CL}$ reduces the possibility that those features of different categories overlap in representation space $R_{anchor}$. Therefore, the introduction of the DTG-Net improves the generalization ability of traditional supervised action recognition network. 

\subsubsection{\textbf{Ablation Study of Supervised DTG-Net}}
\begin{table}[h!]
  \centering
  \tiny
  \setlength{\tabcolsep}{3.2mm}
  \begin{tabular}[]{c|c|c|c|c|c}
       \toprule
       SSL&backbone&DTG&$acc@1$&$acc@5$&$mAP$\\
       \midrule
        UCF&UCF101\\
        \hline
        &TSN&\xmark&63.15&85.54&63.07 \\
        &TSN&\cmark&66.75({\color{red}{+3.6}})&87.02({\color{red}{+1.5}})&66.93({\color{red}{+3.9}}) \\
        \hline
        &HMDB51\\
        \hline
        &TSN&\xmark&30.26&67.32&30.26\\
        &TSN&\cmark&31.96({\color{red}{+1.7}})&67.58({\color{red}{+0.3}})&31.96({\color{red}{+1.7}}) \\
       \hline
        Mini-K200&        UCF101\\
        \hline
        &TSN&\xmark&67.88&88.10&67.95\\
        &TSN&\cmark&68.41({\color{red}{+0.5}})&88.91({\color{red}{+0.8}})&68.45({\color{red}{+0.5}}) \\
        \hline
        &HMDB51\\
        \hline
        &TSN&\xmark&32.55&66.21&32.55\\
        &TSN&\cmark&34.12({\color{red}{+1.6}})&68.69({\color{red}{+2.5}})&34.12({\color{red}{+1.6}}) \\
        \bottomrule
  \end{tabular}
  \vspace{0.5cm}
  \caption{The ablation study results of \textbf{supervised DTG-Net} on the data sets of UCF split1 and HMDB split1. SSL is the data set for the self-supervised task.}
  \label{tab:ablationstudySupervisedDTG}
\end{table}
As we explained above, supervised DTG-Net improves the generalization ability of the action network. We conduct an ablation study to prove that the supervised DTG-Net can improve the supervised action recognition baseline for a large margin. All of these experiments are conducted on both UCF and HMDB datasets. To demonstrate the robustness of the supervised DTG-Net, we initialize the action recognition network with two different pre-trained models of DTG-Net with UCF101 and Mini-Kinetics200. As shown in Table \ref{tab:ablationstudySupervisedDTG}, we compare the supervised DTG-Net and the baseline (typical action recognition network) in four different settings.
 
Table \ref{tab:ablationstudySupervisedDTG} shows that supervised DTG-Net improves $>3\%$  and $>1.5\%$ of top1 accuracy / mAP on UCF101 and HMDB51 separately. This demonstrates that the contrastive self-supervised learning method could alleviate the over-fitting of the action recognition task. We argue that joint learning helps the video action recognition network to remember the initial parameters, which could benefit the action recognition.

\subsubsection{\textbf{Visualization}}
\label{sec:vis_dtgnet}
As we claimed, the supervised DTG-Net improves the generalization of the action recognition networks. Because the adversary learning between the $L_{CL}$ and $L_{CE}$, the overlapping of the features of different categories is alleviated. The feature distribution of supervised DTG-Net is more clear, as shown in the Figure~\ref{fig:visualization_supervised_dtg-net}. 

\subsubsection{\textbf{The Comparison with SOTA}}

\begin{table}[h!]
    \centering
    \small
    \setlength{\tabcolsep}{2.6mm}
    \begin{tabular}[]{c|c|c}
        \toprule
        Methods & UCF & HMDB \\
        \midrule
        DT+MVSM~\cite{DT+MVSMCai2014Multi} & 83.5 & 55.9 \\ 
        iDT+FV~\cite{IDT} & 85.9 & 57.2 \\ 
        TDD+FV &  90.3 & 63.2 \\ 
        C3D+iDT~\cite{C3D}& 90.4 & - \\ 
        LTC+iDT~\cite{C3D} & 92.4 & 67.2 \\
        \hline
        Two Stream~\cite{twostream} & 88.6 & - \\ 
        TSN-BN-Inception~\cite{TSN}& 94.0 & 68.5 \\
        \hline
        C3D~\cite{C3D}&  82.3 & 56.8 \\
        Conv Fusion  & 82.6 & 56.8 \\ 
        ST-ResNet~\cite{ST-NetFeichtenhoferSpatiotemporal} &  93.5 & 66.4 \\ 
        Inception3D~\cite{STCdiba2018spatio} & 87.2 & 56.9 \\ 
        3D ResNet101~\cite{3DRESNEThara2017learning} (16 frames) & 88.9 & 61.7 \\ 
        3D ResNeXt101~\cite{3DRESNEThara2017learning} (16 frames)& 90.7 & 63.8 \\ 
        STC-ResNet101~\cite{STCdiba2018spatio} (16 frames) & 90.1 & 62.6 \\
        DynamoNet~\cite{DynamoNet} (ResNeXt101,32 frames)&91.6&66.2\\
        DynamoNet~\cite{DynamoNet} (ResNeXt101,32 frames)&93.1&68.5\\
        RGB-I3D~\cite{I3D} (ResNet50,32 frames) &94.5&69.0 \\
        \hline
        supervised DTG-Net (ResNet50,16 frames)&95.7&71.8\\
         supervised DTG-Net (ResNet101,8 frames)&96.1&73.3  \\
         \bottomrule
    \end{tabular}
  \vspace{0.5cm}
    \caption{The comparison with the state of the art supervised action recognition methods. }
  
    \label{tab:supervisedSOTA}
\end{table}

Since the supervised DTG-Net is based on a supervised action recognition model, we compare the \textit{supervised DTG-Net} with state of the art supervised action recognition methods. Follow the common practice, and we initialize the action network with the Kinetics400 pre-trained model~\cite{I3D}. This pre-trained model is also used as the teacher in the supervised DTG-Net. The action network (student) in this supervised DTG-Net is an I3D-style 3D-ResNet. 

Figure \ref{tab:supervisedSOTA} shows that the DTG-Net suppresses state of the art supervised action recognition methods with \textit{less input frames}. Also, there is a significant improvement compared with a simple I3D-style 3D ResNet50 action network~\cite{I3D}. 
Therefore, we could say that DTG-Net can improve the state of the art action network further. Because the DTG-Net structure is independent with the action network, its impact on more action networks could be investigated in the future.


\section{Conclusion}
To reduce dependency on large-scale data, the self-supervised learning technique has become an attractive emerging area. SSL has achieved significant results in image-based vision tasks. In this paper, our attention lies in the challenge of video understanding. Specifically, we propose a self-supervised teacher-student architecture, called \textit{DTG-Net}. DTG-Net utilizes two low-cost ways to improve the action recognition tasks with video data, i.e., the unlabeled video data and the prior knowledge from the teacher guidance. 
The teacher guidance speeds up self-supervised learning and reduces the dependency of unlabeled data for SSL. Furthermore, to improve the generalization of the action network, we propose the \textit{supervised DTG-Net}, which introducing the DTG-Net into the supervised action recognition model. Traditional cross-entropy loss in supervised learning suffers the feature overlapping for different categories. Supervised DTG-Net alleviates this problem by the contrastive loss in DTG-Net, which is an adversary with cross-entropy loss. Finally, we compared the proposed method with state-of-the-art methods, which shows that the DTG-Net can achieve competitive results compared with both self-supervised and supervised methods.
{\small
\bibliographystyle{ieee_fullname}
\bibliography{egbib}
}

\end{document}